# idT5: Indonesian Version of Multilingual T5 Transformer

Mukhlish Fuadi
*Dept. of Electrical Engineering*
*Institut Teknologi Sepuluh Nopember*
Surabaya, Indonesia
*Universitas Islam Negeri Maulana Malik Ibrahim*
Malang, Indonesia
muchad@uin-malang.ac.id

Adhi Dharma Wibawa
*Dept. of Electrical Engineering*
*Dept. of Medical Technology*
*Institut Teknologi Sepuluh Nopember*
Surabaya, Indonesia
adhiosa@te.its.ac.id

Surya Sumpeno
*Dept. of Electrical Engineering*
*Dept. of Computer Engineering*
*Institut Teknologi Sepuluh Nopember*
Surabaya, Indonesia
surya@te.its.ac.id

*Abstract*— Indonesian language is spoken by almost 200 million people and is the 10th most spoken language in the world, but it is under-represented in NLP (Natural Language Processing) research. A sparsity of language resources has hampered previous work on Indonesian. The Transformer is a new architecture rapidly becoming dominant for NLP, surpassing alternatives like convolutional and recurrent neural networks. T5 (Text-to-Text Transfer Transformer) is a Transformer model that converts all text-based language problems to text-to-text format for English. The multilingual variant is mT5 (multilingual T5) which has shown promising results on many NLP tasks across languages. However, the size of this multilingual model is a drawback for its application in real production applications, which sometimes require only one language. In this study, the mT5 model was adapted for only one language, Indonesian, resulting in a pre-trained T5 model that was specific only for Indonesian with a smaller size. For performance comparison, we fine-tuned this model and the mT5 model to the Sentiment Analysis (SA), Question Generation (QG), and Question Answering (QA) tasks with the exact mechanism and dataset. Fine-tuned model based on our model achieved 77.18% accuracy on SA, 8% higher than the mT5-based model, and obtained nearly the same score as the mT5-based model on QG and QA. The results confirm that it is possible to produce a smaller pre-trained model that maintains comparable yields while reducing the model size by up to 58%. In addition, the resulting model requires less memory, loads faster, and inference times faster.

*Keywords—Model Compression, Transformer, Pre-trained Model, Question Generation, Question Answering, Sentiment Analysis*

I. INTRODUCTION

Despite being ranked as the 10th most spoken language worldwide [1], Indonesian is still under-represented in NLP (Natural Language Processing) research. Among the obstacles is the limited resources based on Indonesian, including the limited pre-trained language model for Indonesian.

NLP combines linguistic, computational, and statistical studies that produce machines that can understand and respond to language like humans [2]. In NLP, sequence models are prevalent to be used to improve the processing of sequential data, such as written and spoken language, because it allows dealing with sequence characteristics in languages that depend on the relationships between words. Recurrent Neural Network (RNN) is one of them.

RNNs are massively used because they have Long Short Term Memory (LSTM) [3] and Gated Recurrent Unit (GRU) [4] architectures. Both units prevent the gradient vanishing problem by providing a more direct means of backpropagation of gradients. This mechanism helps calculations when the sentence being processed is long. The high flexibility of the network can solve various problems [5]. However, these models are imperfect because their inherent repeating structure makes them difficult to parallelize in many processes and is problematic in long clauses due to missing gradients [6]. Additionally, sequential processing also makes training a very tedious and time-consuming task.

The Transformer is a new architecture that can overcome these problems [7]. Transformers suffer from no loss of gradients and can complete tasks on data sequences while efficiently handling remote dependencies, facilitating the parallelization of tasks, and speeding up training of the more comprehensive network. With its remarkable ability to transfer knowledge from unlabeled data to downstream tasks, pre-trained Transformer-based language models have emerged as an essential component of modern natural language processing systems. Therefore the Transformer is fast becoming the dominant architecture for natural language processing [8].

T5 (Text-to-Text Transfer Transformer) is a Transformer variant that converts all text-based language problems into text-to-text format [9]. This approach is practical because it allows knowledge transfer from high-resource tasks to low-resource tasks without changing the model architecture [10]. The T5 model was initially pre-trained for English but has expanded to a multilingual setup as the mT5 achieves advanced performance on several cross-language comprehension benchmarks [11]. mT5 covers 101 languages, including Indonesian, and has several checkpoints for several pre-trained models. Among them is mT5-base, which has 580 million parameters, so it has a relatively large size.

Multilingual Transformers are a natural language processing model that can handle multiple languages. One of the main advantages of using a multilingual Transformer is that it allows for processing multiple languages using a single model, which can be more efficient and cost-effective than using multiple single-language models. Recent results suggest that multilingual models can perform better than monolingual models, especially for low-resource languages [12]. However, multilingual models are typically trained on a larger amount of data than single-language models. They can be more computationally expensive to train and use.

Some real-world applications only need to handle specific to one language; therefore, we propose reducing the vocabulary size by reducing the number of languages the model handles. In this paper, we extract mT5 to obtain a smaller Transformer model specific to only one language, Indonesian. The contribution of this research is a pre-trained Transformer model for Indonesian. The resulting model can later be fine-tuned for various NLP tasks in Indonesian. This method can also be applied to other languages.

In the next section, we explain the Transformer mT5 model and how we generate an Indonesian-specific model from the multilingual model. Next, we tested the model we produced by comparing its performance with the mT5 model itself. Comparisons were made using the fine-tuning results of the two models on the Question Generation, Question Answering, and Sentiment Analysis tasks. In addition, the comparison is also made to the loading time, inference time, and memory usage. Finally, we analyze and summarize the experimental results at the end of this paper.

## II. RELATED WORK

Several previous studies have been conducted to compress the size of Transformer models. Knowledge distillation [13, 14] is one method that aims to build models with a simpler architecture than the original while mimicking their behavior. This is done by transferring the knowledge learned by the large teacher network to smaller students. This method has been implemented to reduce the number of layers of the BERT model [15] and can effectively compress without reduced performance. Model distillation is not limited to architectural simplification but [16] does so by simultaneously training teachers and students to reduce vocabulary size and embedding size.

The layer pruning strategy proposed by [17] does not require new model training but can be directly applied to previously trained models. This pruning strategy achieves a 40% reduction in model size and a 50% reduction in inference time while retaining up to 98.2% native accuracy across well-known architectures, BERT, XLNet, and RoBERTa. This method encourages a better way to implement the student-teacher architecture because it considers all the lower layers. It does not make students initialize by taking one layer out of every two layers, as did [15].

Knowledge distillation is also applied by [18] by using the distribution of self-attention and value relationships from the teacher's last Transformer layer to guide student training so that it is effective and flexible for student models. In research [19], this method was generalized by using multi-headed self-attention relationships to train students, resulting in finer self-attention knowledge and removing the limit on the number of student attention centers. This study's monolingual and multilingual models distilled from BERT, RoBERTa, and XLM-R obtained competitive performance.

The approach focusing on quantizing model weights is applied by [20] to the Transformer architecture of the BERT model by reducing the model's memory footprint by representing its weights with lower precision values. However, this method is effective when used with specific hardware devices.

Based on analysis and comparison, [21] concluded that traditional model compression methods such as quantization and trimming benefit BERT. Techniques specific to BERT, such as knowledge distillation variants and methods that reduce the number of architectural hyperparameters, also yield competitive results.

Regarding the multilingual Transformer model, [22] reduces the size of the mBERT vocabulary, thereby reducing the number of parameters primarily located in the embeddings layer. This method reduced up to 45% of the total parameters without reducing the average accuracy. Another advantage of this method is that there is no need to train the model from scratch like [23]. A similar idea was also used by [24] to extract the Russian language from mT5 but has yet to have a peer-reviewed publication. The compression is done by only keeping the tokens used frequently and trimming the others. This paper uses the same method to derive a smaller and more specific model for Indonesian.

## III. METHODOLOGY

Multilingual T5 (mT5) is a multilingual variant of T5 which has been pre-trained with the mC4 (Multilingual Colossal Clean Crawled Corpus) dataset, which includes 101 languages, including Indonesian [11]. This model has a reasonably large size because it contains a variety of languages. Based on the mT5 model provided by the Google AI Team and published on the Hugging Face repository website, the mT5-base version has a size of 2.33 GB. The smallest model, mT5-small, has a size of 1.2 GB.

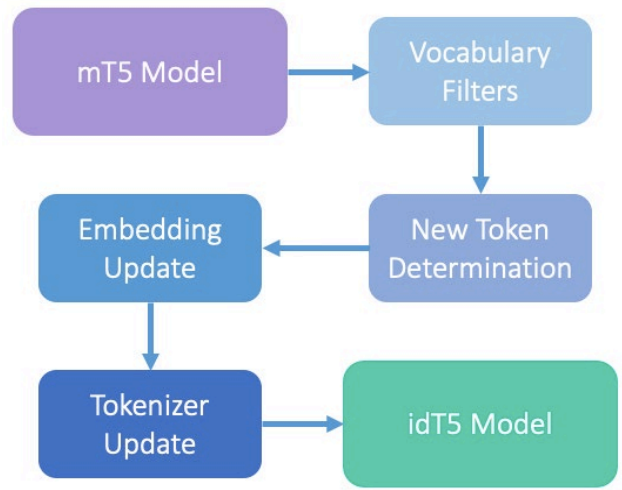

Fig. 1. Flow of extraction.

To obtain the Indonesian language version of mT5, we extract the Indonesian language from the mT5-base by selecting Indonesian vocabulary and then updating the embedding layer and tokenizer to produce a suitable model. The detailed flow of the stages is shown in Fig. 1.

### A. Selecting Vocabularies

We use the original tokenizer to process the Indonesian language corpus. Since English is often used in Indonesian texts, we maintain a small number of English tokens in the model. To calculate the frequency of different tokens, we retrieved the corpus of Indonesian and English sentences from the Leipzig corpora collection [25].

**Vocabulary Filters**. Based on the token count in the Indonesian corpus, we found that only 29% of the model vocabulary was used and 27% of English. Interestingly, there is over 76% overlap between the vocabularies. This is possible because in Indonesian texts, sometimes there are English words or other words with Latin letters. Moreover, the top 20K tokens constitute over 98% of the Indonesian corpus. For English, the statistics are similar.

**New Token Determination**. Based on vocabulary filters, in this study, the number of tokens is determined at 30K with the composition of 1K of top tokens from the original tokenizer, 10K from the English token, which already covers more than 95% of the English corpus, and 100 special tokens used by T5. As for Indonesian tokens, the remaining tokens have yet to be accommodated in the three token groups, so the total number of tokens is 30K or 12% of 250K tokens in the multilingual version. As for the other tokens removed. Based on the 30K tokens, the Indonesian language tokens accommodated are 24,471. This can happen because there are overlapping tokens.

### B. Updating The Model

**Embedding Update**. To update the neural network, we replaced its input and output parameters with our predefined parameters, thereby reducing the model size by 58%, from 2330 MB to 977 MB. The new model has 244 million parameters, or 42% of the multilingual model.

**Tokenizer Update**. To update the tokenizer, we use the Protobuf representation as T5 uses the Sentencepiece tokenizer.

The resulting model has been uploaded to Transformers Hub to make it easier for NLP communities to use, especially those related to the Indonesian language[1]. This model is a pre-trained model like the mT5 model. Therefore it needs to be fine-tuned to complete NLP tasks because mT5 is trained only on unsupervised tasks to predict missing words.

A pre-trained model is a model that has already been trained on a large dataset and can be used for various natural language tasks without the need for further training [26]. Pre-trained models are useful because training a model from scratch can be time-consuming and reduce computational costs. While fine-tuning is a technique used in NLP to adapt a pre-trained model to a specific task [27]. In fine-tuning, a pre-trained model is further trained on a new dataset to improve its performance on a specific task. This is often more efficient than training a model from scratch.

## IV. RESULTS AND DISCUSSION

The model we produce (idT5) is an adaptation of mT5, so it is also a pre-trained model like mT5, which must be fine-tuned to be able to complete NLP tasks. The idT5 model is a smaller version of the mT5, so it is possible that its performance may be worse than the original mT5 version. We present a simple comparison in this section to find out the performance of idT5 compared to mT5 when fine-tuned to NLP tasks. We fine-tuned several NLP tasks on both models: Question Generation (QG), Question Answering (QA), and Sentiment Analysis. In order to do a fair and apple-to-apple performance comparison, we use the exact mechanisms, datasets, and configurations in each model. We only used one epoch for fine-tuning in all tasks because it was only to compare the performance of two models under the same conditions, not to find the best performance. All fine-tuning and comparisons were conducted using the Hugging Face Transformers Trainer class [8] with PyTorch [28] on Google Colab.

### A. Question Generation

Question Generation is a NLP task in which a model is trained to generate questions from a given input text automatically [29]. This can be useful for various applications, such as improving reading comprehension or creating question-answering systems. Question Generation is one of the most challenging jobs in NLP [30-32]. Among the difficulties is that the system must be able to recognize text well so that questions can be well structured, relevant, and answerable [33].

In the Question Generation and Question Answering tasks, to generate a model based on idT5 and a model based on mT5, we used the same method used by [34]. For the dataset, we used the TyDi QA gold passage task dataset, a dataset about reading comprehension [35]. The dataset includes 11 languages, but we only use the Indonesian language dataset. The training set consists of 5702 Indonesian-language contexts and the same number of questions. As for the dev set, there are 565 contexts and the same number of questions. We fine-tuned the Question Generation and Question Answering tasks using the TyDi QA training set and dev set with hyperparameters: learning rate 1e-4 and training epoch 1.

TABLE I. GENERATED QUESTION FROM TYDIQA DEV SET

| | 1 |
|---|---|
| Context & Answer | Kaktus adalah nama yang diberikan untuk anggota tumbuhan berbunga famili Cactaceae. <br><br>*(Cactus is the name given to members of the flowering plant family Cactaceae.)* |
| Target Question | Apa nama ilmiah tumbuhan kaktus ? <br>*(What is the scientific name of the cactus plant?)* |
| mT5 | Apa yang dimaksud dengan Kaktus? <br>*(What is meant by Cactus?)* |
| idT5 | Apa itu Kaktus? <br>*(What is Cactus?)* |
| | 2 |
| Context & Answer | Virusnya sendiri bernama Human Immunodeficiency Virus (atau disingkat HIV) yaitu virus yang memperlemah kekebalan pada tubuh manusia. Orang yang terkena virus ini akan menjadi rentan terhadap infeksi oportunistik ataupun mudah terkena tumor. Meskipun penanganan yang telah ada dapat memperlambat laju perkembangan virus, namun penyakit ini belum benar-benar bisa disembuhkan. <br><br>*(The virus itself is called Human Immunodeficiency Virus (or HIV for short), which is a virus that weakens immunity in the human body. People who are exposed to this virus will be susceptible to opportunistic infections or prone to tumors. Even though existing treatments can slow the rate of development of the virus, this disease has not been completely cured.)* |
| Target Question | Apakah kepanjangan dari HIV ? <br>*(What does HIV stand for?)* |
| mT5 | Apa yang dimaksud dengan HIV? <br>*(What is meant by HIV?)* |
| idT5 | Apa nama virus yang memperlemah kekebalan pada tubuh manusia? <br>*(What is the name of a virus that weakens immunity in the human body?)* |

---
[1] https://huggingface.co/muchad/idt5-base

To compare the two fine-tuned models, we evaluated them with the TyDi QA dev set. To quantify the performance of these two models, we used the evaluation package by [36] which calculates BLEU (Bilingual Evaluation Understudy) [37] and ROUGE (Recall-Oriented Understudy for Gisting Evaluation) [38]. BLEU measures precision by counting how many words in machine-generated questions appear in human reference questions. Rouge measures recall, how many words in a human reference question appear in a machine-generated question. Samples of generated questions can be seen in TABLE I.

TABLE II. PERFORMANCE COMPARISON ON QUESTION GENERATION

| Model | BLEU-1 | BLEU-2 | BLEU-3 | BLEU-4 | ROUGE-L |
|---|---|---|---|---|---|
| mT5 | **34.39** | 26.61 | 21.32 | 17.25 | **36.63** |
| idT5 | 34.35 | **27.08** | **21.79** | **17.71** | 36.40 |

The performance comparison results for the Question Generation task are shown in TABLE II. When comparing the scores using BLEU and ROUGE-L metrics, the two fine-tuned models (mT5-based and idT5-based) showed almost identical scores, with only a minimal difference. The mT5-based model performed better on BLEU-1 and ROUGE-L, while the idT5-based model performed better on BLEU-2, BLEU-3, and BLEU-4. Overall, these results indicate that the idT5-based model achieved similar performance to the mT5-based model on the Question Generation task, despite having a smaller size.

### B. Question Answering

Question Answering systems are designed to automatically answer questions posed in natural language by humans. These systems can use pre-structured databases or natural language document collections to generate answers [39]. Question Answering systems are commonly used to create intelligent assistants or chatbots that can understand and respond to user queries [40]. Question Answering systems often employ a combination of natural language understanding, information retrieval, and knowledge representation techniques to provide accurate answers.

TABLE III. PERFORMANCE COMPARISON ON QUESTION ANSWERING

| Model | EM | F1 |
|---|---|---|
| mT5 | **60.88** | **74.92** |
| idT5 | 60.53 | 73.64 |

In the Question Answering task, we evaluated the answers generated by the idT5 and mT5 based models with the answers from the TyDi QA dev set as a reference using F1 and exact match scores. The exact match (EM) metric measures the percentage of predictions that exactly match a ground truth answer. In contrast, the F1 score metric assesses the average overlap between the prediction and the ground truth answer [41]. The comparison of the scores of the mT5 and idT5 based Question Answering models is shown in TABLE III. In this task, the mT5-based model was superior to the idT5-based model but with a slight difference for both exact match and F1 scores. TABLE IV shows sample answers from both models.

TABLE IV. GENERATED ANSWER FROM TYDIQA DEV SET

| | 1 |
|---|---|
| Context & Target Answer | Rumah Baanjung (Ba'anjung) adalah nama kolektif untuk rumah tradisional suku Banjar dan suku Dayak Bakumpai.[1] Suku Banjar biasanya menamakan rumah tradisonalnya dengan sebutan Rumah Banjar atau Rumah Bahari.<br><br>*(Rumah Baanjung (Ba'anjung) is the collective name for the traditional houses of the Banjar and Dayak Bakumpai tribes.[1] The Banjar tribe usually names their traditional houses as Banjar Houses or Maritime Houses.)* |
| Question | Apakah nama rumah adat suku Banjar? *(What is the name of the traditional Banjar house?)* |
| mT5 | Baanjung *(Baanjung)* |
| idT5 | Rumah Bahari *(Maritime Houses)* |
| | 2 |
| Context & Target Answer | BlackBerry pertama kali diperkenalkan di Indonesia pada pertengahan Desember 2004 oleh operator Indosat dan perusahaan Starhub. Perusahaan Starhub merupakan pengejewantahan dari RIM yang merupakan rekan utama BlackBerry. Pasar BlackBerry kemudian diramaikan oleh dua operator besar lainnya di tanah air yakni Excelcom dan Telkomsel.[2]<br><br>*(BlackBerry was first introduced in Indonesia in mid-December 2004 by Indosat operator and Starhub company. Starhub company is the embodiment of RIM which is BlackBerry's main partner. The BlackBerry market was then enlivened by two other major operators in the country, namely Excelcom and Telkomsel.[2])* |
| Question | Kapan BlackBerry mulai masuk ke Indonesia ? *(When will BlackBerry start entering Indonesia?)* |
| mT5 | Desember 2004 *(December 2004)* |
| idT5 | pertengahan Desember 2004 *(mid-December 2004)* |

### C. Sentiment Analysis

Sentiment Analysis is a kind of text classification that catalogs texts based on the sentiment orientation, such as opinions or emotions [42]. This can be useful for various applications, such as identifying the sentiment of customer reviews or detecting the emotional tone of social media posts. Sentiment Analysis is often performed using machine learning algorithms trained on large datasets labeled with the corresponding sentiment of each piece of text. These algorithms can then automatically classify new text according to its sentiment.

In this task we used the SmSA dataset [43], a collection of comments and reviews from various online platforms in Indonesia. The dataset has three possible sentiments: positive, negative, and neutral. However, in this evaluation, we only used positive and negative data. So we get a training set of 9852 out of a total of 11000 sentences, a validation set of 1129 out of 1260, and a test set of 412 out of 500. For fine-tuning Sentiment Analysis tasks on idT5 and mT5 models, we use the hyperparameters: learning rate 3e-4 and training epoch 1.

TABLE V. PERFORMANCE COMPARISON ON SENTIMENT ANALYSIS

| Model | Accuracy | F1 |
|---|---|---|
| mT5 | 68.93 | 67.51 |
| idT5 | **77.18** | **77.01** |

To compare the performance of the models on the Sentiment Analysis task, we used accuracy and F1 macro-average scores. TABLE V indicates that the idT5-based model outperformed the mT5-based model on both metrics. The idT5-based model achieved 8% higher than the mT5-based model on accuracy and F1 scores. These results suggest that the idT5-based model is superior to the mT5-based model on the Sentiment Analysis task.

Based on the comparison of the performance of the three tasks, the results indicate that the idT5-based model performed similarly to the mT5-based model on the question generation task but outperformed the mT5-based model on the sentiment analysis task. The mT5-based model performed slightly better than the idT5-based model on the question-answering task. Overall, the results suggest that the idT5-based model, which is 58% smaller than the mT5 model, can achieve similar or better performance than the mT5-based model on specific natural language processing tasks.

We are interested in producing a specific model for one language because its implementation sometimes only requires tasks focusing on one language. Models for only one language are smaller and save more memory than multilingual Transformer models. In this paper, we reduce the parameter number of multilingual Transformer models, resulting in smaller models that require less memory. Memory limits are critical, especially when deploying Transformers on public cloud platforms. In addition, the smaller models load faster than the larger ones, which can also increase the speed with which applications are deployed. The method we use can also reduce the inference time.

TABLE VI. OVERHEAD COMPARISON

| Model | Loading (Sec) | Sequence Length = 128 | | Sequence Length = 512 | |
| --- | --- | --- | --- | --- | --- |
| | | Memory (MB) | Inference (Sec) | Memory (MB) | Inference (Sec) |
| mT5 | 12.11 | 4322 | 0.239 | 8094 | 1.103 |
| idT5 | **7.82** | **2170** | **0.158** | **3436** | **0.784** |

TABLE VI shows both models' loading time, memory allocation, and inference time. The loading time measurement was carried out on Google Colab (2vCPU @ 2.2GHz, 13GB RAM) by taking the average value of 10 tests on each model. The idT5 model has a smaller model size with fewer parameters, so the loading time is faster than the mT5 model.

Measurement of memory allocation and inference time is done with batch size = 8 on Google Colab with GPU (Tesla T4 – 16GB). The tests were carried out with a sequence length = 128 and a sequence length = 512. At both sequence lengths, idT5 requires a smaller memory allocation, even less than half of the memory required by the mT5 model. In the mT5 model, a sequence length of 512 requires 46.6% more memory than the memory required for a sequence length of 128. Meanwhile, for idT5, the difference in memory required from sequence length 128 to sequence length 512 is only 36.8%.

The inference time was measured as the average of 6 tests. Although this method does not change the model architecture, it still leads to faster inference times compared to the method proposed in [22], which does not improve the inference time.

Based on this test, the specific language model, idT5, has a faster loading time, requires less memory allocation, and has a faster inference time than a comparable multilingual model, mT5. These results suggest that specific language models can have performance advantages over multilingual models in certain scenarios. A model like this will be helpful when deploying the models for specific language needs and on public cloud platforms or when memory constraints are an issue.

V. CONCLUSION AND FUTURE WORK

One advantage of using a multilingual Transformer is that it can handle text in multiple languages without needing a specific model for each language. This can be useful in applications where processing text in multiple languages is necessary. However, the multilingual model is less efficient for tasks requiring only one language. Its large size will affect memory allocation and processing speed. Memory limitations will be a consideration when deploying Transformers on public cloud platforms.

This paper presents a simple method for creating targeted single-language models from multilingual Transformer models. This method allows us to produce models specific to a particular language with smaller sizes than the original multilingual model. We demonstrate this method using the mT5-base Transformer model and extract Indonesian and a small portion of English since English vocabulary is commonly used in Indonesian texts. The resulting model is 58% smaller than the original multilingual model, demonstrating this method's potential for creating more compact models tailored to specific languages.

In this study, we fine-tuned the smaller and mT5 models on three natural language processing tasks (Question Generation, Question Answering, and Sentiment Analysis) to compare their performance. We found that the smaller model performed similarly to the mT5 model on these tasks. Additionally, the smaller model required less memory, had faster loading times, and had faster inference times, even though the model architecture was not changed. These results indicate the potential of this method for creating smaller, more efficient single-language models from multilingual models.

As future work, it would be interesting to evaluate this method on a wider range of models and tasks to see if the results are consistent. Testing this method on other languages would also be useful for creating specific-language models for various tasks. This model has the potential to facilitate the use of the Indonesian Transformer in real-world applications, and we plan to continue exploring its potential in the future.